\documentclass[letterpaper, 10 pt, conference]{ieeeconf}  

\IEEEoverridecommandlockouts                              

\usepackage{cite}
\usepackage{comment}
\usepackage{graphicx}
\usepackage{graphics}
\usepackage{float}
\usepackage{graphicx}
\usepackage{graphics}
\usepackage{mdwlist}
\usepackage{xcolor}
\usepackage{fancyvrb}
\usepackage{amssymb}
\usepackage{amsmath}
\usepackage{amsfonts}
\usepackage{amssymb}
\usepackage{array}
\usepackage{subfigure}
\usepackage{ragged2e}
\usepackage{tabu}
\usepackage{balance}
\usepackage{algorithmicx}
\usepackage{algorithm}
\usepackage{algpseudocode}
\usepackage[marginal]{footmisc}
\usepackage{graphicx}
\usepackage{subfigure}
\usepackage{float}
\usepackage{balance}
\usepackage{epstopdf}

\usepackage{array,makecell}
\usepackage{microtype}
\usepackage{graphicx}
\usepackage{subfigure}
\usepackage{booktabs} 
\usepackage{hyperref}
\usepackage{amsmath}
\usepackage{cleveref}
\title{\LARGE \bf
Robust Reinforcement Learning via Genetic Curriculum
}

\author{Yeeho Song$^{1}$ and Jeff Schneider$^{1}$

\thanks{$^{1}$Robotics
    Institute, Carnegie Mellon University, Pittsburgh, PA,
    USA.  \texttt{\{yeehos, jeff4, \}@andrew.cmu.edu}}%
\thanks{This work was partly funded by The Boeing Company,}
}

\begin{document}

\maketitle
\thispagestyle{empty}
\pagestyle{empty}

\begin{abstract}
Achieving robust performance is crucial when applying deep reinforcement learning (RL) in safety critical systems. Some of the state of the art approaches try to address the problem with adversarial agents, but these agents often require expert supervision to fine tune and prevent the adversary from becoming too challenging to the trainee agent. While other approaches involve automatically adjusting environment setups during training, they have been limited to simple environments where low-dimensional encodings can be used. Inspired by these approaches, we propose genetic curriculum, an algorithm that automatically identifies scenarios in which the agent currently fails and generates an associated curriculum to help the agent learn to solve the scenarios and acquire more robust behaviors. As a non-parametric optimizer, our approach uses a raw, non-fixed encoding of scenarios, reducing the need for expert supervision and allowing our algorithm to adapt to the changing performance of the agent. Our empirical studies show improvement in robustness over the existing state of the art algorithms, providing training curricula that result in agents being 2 - 8x times less likely to fail without sacrificing cumulative reward. We include an ablation study and share insights on why our algorithm outperforms prior approaches.


\end{abstract}

\section{Introduction}

When training an RL agent, learning to solve the remaining 10\% of the scenarios is often significantly more difficult compared to learning to solve the first 90\% of the scenarios. This presents a challenge to using RL in safety critical applications, such as autonomous vehicles, where robustness, the probability of an agent not landing in irreversible and catastrophic states (i.e. collision) plays a crucial role in determining product viability. In a typical RL setup, as an agent's performance improves, it becomes not only rare to encounter and collect data on the scenarios in which the agent does poorly but also difficult to learn new behaviors when approaching a local minimum. This results in an agent converging to a suboptimum with several scenarios left unsolved.

One prominent approach for robust RL is to use adversarial agents to inject adversarial noise to explore challenging situations. However, adversarial agents often converge to the worst case scenario in which the protagonist cannot learn and requires expert supervision to avoid this issue. Some scenarios are not well represented by adversarial noise, such as a particular sequence of tasks or environment setup difficult for the agent. While other approaches involve encoding the environment or generating a curriculum to help learn difficult tasks, they are mostly limited to benchmarks with small scenario space where low-dimensional encodings can be used.

In this paper, we propose genetic curriculum (GC) which uses a genetic algorithm to generate curricula for training robust RL agents. By running a genetic algorithm, GC will generate training scenarios that the agent cannot solve, helping the agent to explore the scenario space efficiently. As scenarios generated by the genetic algorithm will be similar to each other, a skill learned from one scenario can easily be transferred to another scenario, allowing them to work as a curriculum helping the agent to learn faster and converge to a more optimal policy. As our algorithm is non-parametric, it can use raw scenario encoding of non-fixed length, minimizing expert supervision of designing encoding methods and helping support highly complex scenario description as the agent's performance improves. 

\section{Related Works}
In robust RL where an agent should be trained against and verified in a variety of different scenarios, recent advances in sim2real ~\cite{akkaya2019solving, chebotar2019closing, kaushik2020fast, hansen2020self} and high fidelity simulators ~\cite{Dosovitskiy17, laminar2017} makes it feasible to collect realistic training data in scenarios too dangerous and difficult to collect in real life. However, even with this setup, an agent would often leave a long tail of unsolved scenarios. As an agent becomes more robust, it becomes less likely to encounter and collect data from situations where the agent fails. Even when data is available, it is often difficult to learn new skills as the agent would often be approaching a local minimum optimized towards more probable scenarios.

Adversarial training is one such method for gathering data in the region where the agent does not do well. Showing success with classical RL ~\cite{iyengar2005robust, nilim2005robust, mannor2012lightning} and deep learning architectures ~\cite{goodfellow2014generative, kurakin2018adversarial, samangouei2018defense, xiao2018generating}, adversarial training in RL pairs a protagonist agent with an adversary agent each playing a zero-sum game of maximizing/minimizing reward in the environment. Robust adversarial RL (RARL)~\cite{pinto2017robust} uses an adversary to apply external force to the protagonist. Risk averse robust adversarial RL (RARARL)~\cite{pan2019risk}, probabilistic action robust Markov decision process (MDP) and robust action robust MDP (PRMDP / NRMDP) ~\cite{tessler2019action} uses adversaries to inject action noise. However, some challenging situations are difficult to represent as noise, such as particularly hard scenarios or environment setups. Also, such a min-max setup often leads to the adversary quickly converging to a worst-case scenario too difficult for the protagonist to learn. Our approach differs by not only encoding scenarios and environment setup instead of adversarial noise but also generating supporting scenarios that help the agent to learn new skills.

Fingerprint policy optimization (FPO) ~\cite{paul2018fingerprint} shares insights on encoding environments and scenarios. Building upon the previous works on classical RL ~\cite{ciosek2017offer} ~\cite{paul2018alternating}, FPO uses Bayesian optimization to select the training setup with the biggest expected performance improvement. However, such approaches have been limited to low-dimensional fixed-length encoding for training environments. Our approach differs by using non-parametric optimizers to use non-fixed length encoding. This allows us to minimize expert supervision by directly using the raw values of a simulator while being more versatile to adapt to the agent's changing needs with no information loss.

Curricular learning explores generating supporting scenarios to help learn new skills. Organizing training data to gradually introduce more complex concepts, ~\cite{bengio2009curriculum}, curricular learning has shown success in supervised learning tasks ~\cite{zaremba2014learning, bengio2015scheduled, graves2017automated, kumar2010self, jiang2015self} as well as various RL tasks ~\cite{asada1996purposive, karpathy2012curriculum,held2018automatic, ivanovic2019barc, florensa2017reverse, florensa2017automatic, narvekar2019learning, fournieraccuracy}. Automatically generating a curriculum of similar yet gradually more complex scenarios is an ongoing question in curriculum learning. Self-paced deep RL (SPDL) ~\cite{klink2020self} is one such approach of exploring how curriculum can automatically be generated based on the agent's current performance. At each epoch, SPDL locates the distribution of scenarios the agent currently performs well and will select training scenarios as a distribution progressively moving towards the goal distribution. However, SPDL likewise has been limited to low-dimensional fixed length encoding of scenarios. We will explore non-parametric approaches to expand the ideas to non-fixed raw encoding of curriculum.

\vspace{-0.4cm}
\begin{figure}[ht]
\begin{center}
\centerline{\includegraphics[width=\columnwidth]{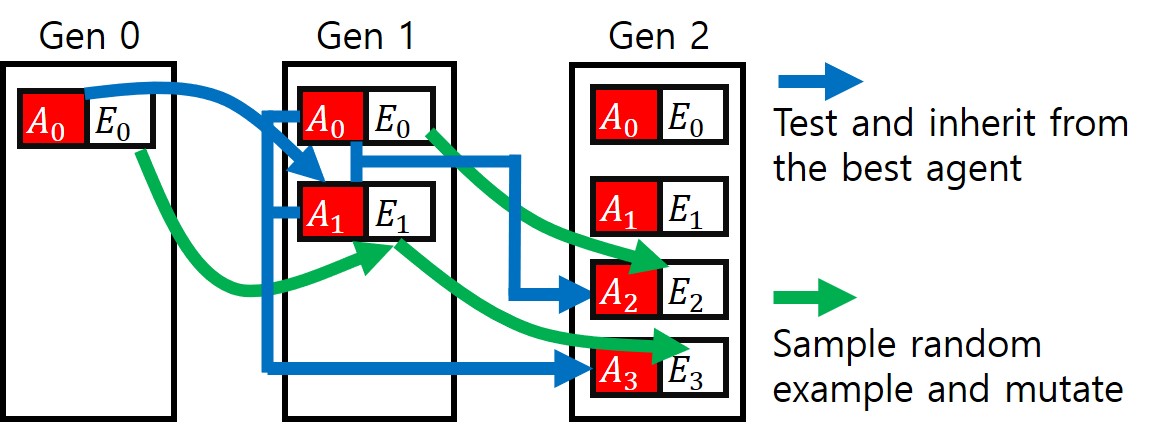}}
\caption{Curriculum generation during POET\cite{wang2019paired}. Unlike our proposed approach, POET runs training inside genetic algorithm, greatly increasing computational load.}

\label{fig:POET}
\end{center}
\vskip -0.3in
\end{figure}

Paired open-ended trailblazer (POET) ~\cite{wang2019paired} borrows several elements from genetic algorithms to use non-fixed length encodings. As shown in ~\Cref{fig:POET}, POET starts by adding a pair consisting of a random agent $(A_0)$ and a random training example $(E_0)$. During an epoch, all pairs that haven't reached a satisfactory performance are trained in parallel. At the end of each epoch, a new pair is generated by adding a small random perturbation, or mutation, to an existing example and pairing it up with a new agent that inherits the parameter weights of the existing agent that performs best on the new example.  As POET has training nested inside the genetic algorithm, the algorithm is computationally expensive as resources spent by any agents not contributing to the performance of the best-performing agent are wasted. Our approach is computationally more efficient by separately running the genetic algorithm outside of the loop. At the end of each epoch, the current policy is fixed and the genetic algorithm runs to generate a set of scenarios the agent cannot solve. Furthermore, POET uses mutation to generate new scenarios. This makes the search for new scenarios an inefficient random walk and makes it difficult to advance towards scenarios drastically different from the starting scenario. Our approach on the other hand incorporates crossover, the processing of mixing sequences from two parent sequences to generate two offspring, to help cover a wider variety of scenarios faster.


\section{Background}

This paper examines continuous space MDP represented as a tuple: [$S,A,P_\psi,r_\psi,\gamma$] where $S$ is a set of states, $A$ action space, and $\gamma \in [0,1)$ is the temporal discount factor. Scenario $\psi$ are not fully observable to the agent, such as obstacles situated out of the line of sight or an internal systems failure. $P_\psi$ and $r_\psi$ represent the state transition dynamics and reward function of the scenario. In the event of a partial engine failure related to fuel pumps, the engine would be burning less fuel per second and will be delivering less thrust, hence the state transition probability and reward on fuel usage will be different from those of a nominal scenario. The agent's policy, $\pi(a\mid s)$ maps states $s \in S$ to $a \in A$. The utility of a policy $\pi$ for scenario $\psi$ is the expected return, $J_{\psi}(\pi) = \mathbb{E}_{a_t\sim\pi}\sum_{t} \gamma^t r_{\psi}(s_t,a_t)$. 

During training, an RL algorithm seeks the optimal policy $\pi^*$ by exploring and gathering data about reward and state dynamics. The data gathered will be dependent on the distribution of scenarios the agent experienced during training $p_{train}(\psi)$;

\vspace{-2mm}
\begin{equation}
\label{eq:RL}
  \pi^* = argmax \sum_{\psi}p_{train}(\psi)J_{\psi}(\pi)
\end{equation}
\vspace{-2mm}

\section{Problem Statement}

For a given scenario, $\psi$, we measure success as follows;

\vspace{-2mm}
\begin{equation}
\label{eq:failure}
  G_{\psi}(\pi)=\left\{
  \begin{array}{@{}ll@{}}
    0, & \text{if}\ \pi \text{ fails for } \psi \\
    1, & \text{otherwise}
  \end{array}\right.
\end{equation}
\vspace{-2mm}

Failure is defined as failing to achieve a goal before exhausting a resource budget set by the user. This could be a walking robot falling down before reaching a target, a wheeled robot crashing into a stationary object, a manipulator robot reaching certain time steps with a cumulative reward lower than a threshold. A robust algorithm should minimize the probability of failure during testing;

\vspace{-2mm}
\begin{equation}
\label{eq:robust}
  \pi_{robust}^* = argmax \sum_{\psi} p_{test}(\psi)G_{\psi}(\pi)
\end{equation}
\vspace{-2mm}

Ideally, the trained agent should satisfy the robustness criteria $\pi^{*}=\pi^{*}_{robust}$. As the distribution of scenarios encountered during testing, $p_{test}({\psi})$, and the definition of failure $G_{\psi}(\pi)$, are problem specific, most papers focus on $J_{\psi}(\pi)$ and $p_{train}({\psi})$. While reward shaping with $J_{\psi}(\pi)$ is possible, such as giving a high penalty towards failure, this often requires expert supervision and fine-tuning for the training to be stable. We, therefore, take the curricular approach of investigating how $p_{train}({\psi})$ can be better selected to train a robust agent.

\section{Approach}

To train a robust agent, we select training scenarios as scenarios the agent currently fails in. Solving these examples directly addresses ~\Cref{eq:robust}. We also select the scenarios to be similar to each other. This follows the idea of curricular learning on building a set of similar scenarios with varying types of challenges and levels of difficulty to help transfer skills from one task to another more easily. Also, just as adversarial RL adds perturbations to make an agent robust to a variety of situations, the differences in our scenarios will help an agent learn not only a specific task but also a variety of similar tasks as well. This paper proposes GC, which borrows concepts from genetic algorithms and curriculum learning to achieve these goals.

At each epoch, curriculum generation starts by initializing a population $\Psi_{population}$ of size $M_{pop}$ with randomly generated scenarios. We express scenarios as a sequence of none-fixed length $\psi = (z_0,z_1,z_2,...)$ where each vector $z$ defines the order of values to be used which would otherwise be filled in by a random number generator in the original benchmark. Factors of variation include size and duration of obstacles and bumps on terrain to the time of occurrence, or type and magnitude of an actuator failure of a legged robot depending on the benchmark. At each iteration, $\Psi_{population}$ is evaluated by current policy $\pi$. $\psi$ is appended to $\Psi_{training}$ if $\pi$ is unable to solve $\psi$.

\begin{figure}[ht]
\vskip 0.2in
\begin{center}
\centerline{\includegraphics[width=\columnwidth]{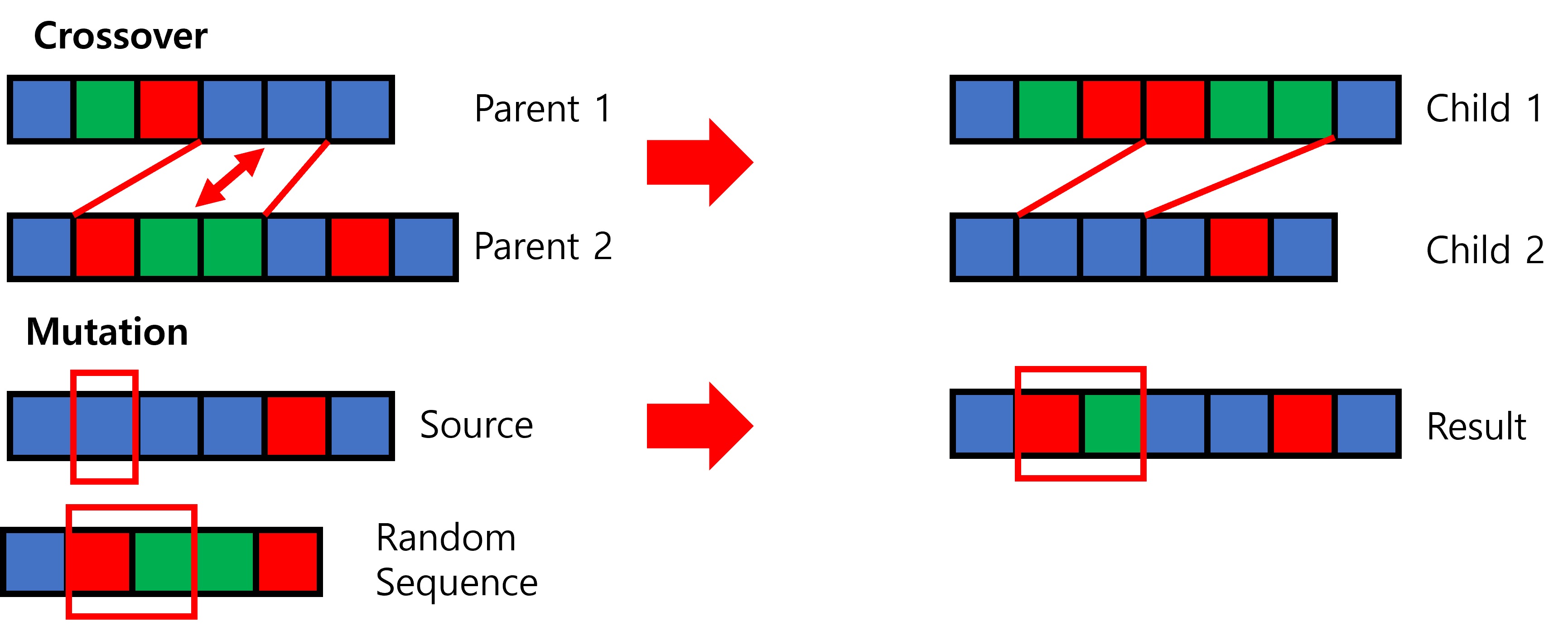}}
\caption{Visualization on crossover and mutation. Our approach use genetic algorithm to use raw, non-fixed length encoding to generate similar scenarios that can act as a curriculum and dynamically change length of encoding to keep up with agent's performance.}
\label{fig:cross_and_mutate}
\end{center}
\vskip -0.3in
\end{figure}

The next $\Psi_{population}$ is generated by crossover. With $L(\psi)$ as the length of encoding for $\psi$, the probability of a scenario being chosen as a parent for a crossover operation is higher if the scenario's encoding is shorter. This encourages sequences to only retain the sections critical to failure and avoid having offspring diverse in irrelevant ways. A selected parent change a random section of its encoding with a random section of another parent's encoding as shown in ~\Cref{fig:cross_and_mutate}. The crossover operation repeats until $|\Psi_{population}| \geq M_{pop}$. To introduce new vectors to the gene pool, every $\psi$ in $\Psi_{population}$ has mutation probability $p_{\mu}$. The mutation is equivalent to conducting crossover with a randomly generated sequence as shown in ~\Cref{fig:cross_and_mutate}.

Once $|\Psi_{population}| \geq M_{pop}$, GC exits the scenario generation cycle and $\Psi_{training}$ is used to train $\pi$ for an epoch. ~\Cref{eq:pseudocodes} shows the pseudocode of our proposed approach.

\vspace{-0.8cm}

\vspace{-3mm}
\begin{equation}
\label{eq:fitness}
  p_{parent}(\psi)=\left\{
  \begin{array}{@{}ll@{}}
    \frac{1}{(max(L(i))-L(i)+1)}, & \text{if}\ \pi \text{ fails } \psi \\
    0, & \text{otherwise}
  \end{array}\right.
\end{equation}
\vspace{-3mm}

\begin{algorithm}[H]  \small
	\caption{Genetic Curriculum (GC)}
	\begin{algorithmic}[1]
		\State \textbf{Initialize} Policy $\pi_0$
		\State \textbf{Input} training steps, iterations, epochs, $M_{train}$, $M_{pop}$, $p_{\mu}$
		\For{$i$ in $epochs$}
    		
    		\State \textbf{Initialize} $\Psi_{population}$
    		\State \textbf{Initialize} $\Psi_{train}=\{\}$
    		\For{$k$ in $\text{iterations}$}
    		    \State Fitness = evaluate($\Psi_{population},\pi_{i+1}$)
    		    \State $\Psi_{train}$ = collect($\Psi_{train}$,$\Psi_{population}$,Fitness)
    		    \If{$|\Psi_{train}|>M_{train}$}
    		    \State Break
    		    \EndIf
    		    \State $\Psi_{population}$ = crossover($\Psi_{population}$,$M_{pop}$,Fitness)
    		    \State $\Psi_{population}$ = mutate($\Psi_{population}$,$p_{\mu}$)
    		\EndFor
    		
    		\While{steps $<$ training steps}
    		    \State $\pi_{i+1}$ = \text{Train Agent}($\pi_i$, $\Psi_{train}$)
    		\EndWhile
    		
		\EndFor
	\end{algorithmic}
	\label{eq:pseudocodes}
\end{algorithm}
\vspace{-3mm}
Our algorithm has several desirable features. As a non-parametric optimizer, it is easy to adapt to various types of tasks and policies. $\psi$ not only has no fixed length, but as visualized in ~\Cref{fig:cross_and_mutate}, offspring scenarios can easily get longer or shorter during curriculum generation. This allows encoding length to expand and contract as needed to accommodate changes in the agent's performance over time. During the experiments, the encoding dimension dynamically changed from 20 - 300D, which would have been difficult with Bayesian optimization as used in FPO ~\cite{paul2018fingerprint}. Another feature visible in ~\Cref{fig:cross_and_mutate} is that scenarios within a curriculum will be similar to each other. With the crossover and mutation operations, all scenarios have part of their sequence shared recurring in another scenario. This similarity makes it easier to transfer skills from one to another.


\section{Experiments}

\subsection{Benchmarks}

\vspace{-0.8cm}
\begin{figure}[ht]
\vskip 0.2in
\begin{center}
\centerline{\includegraphics[width=\columnwidth]{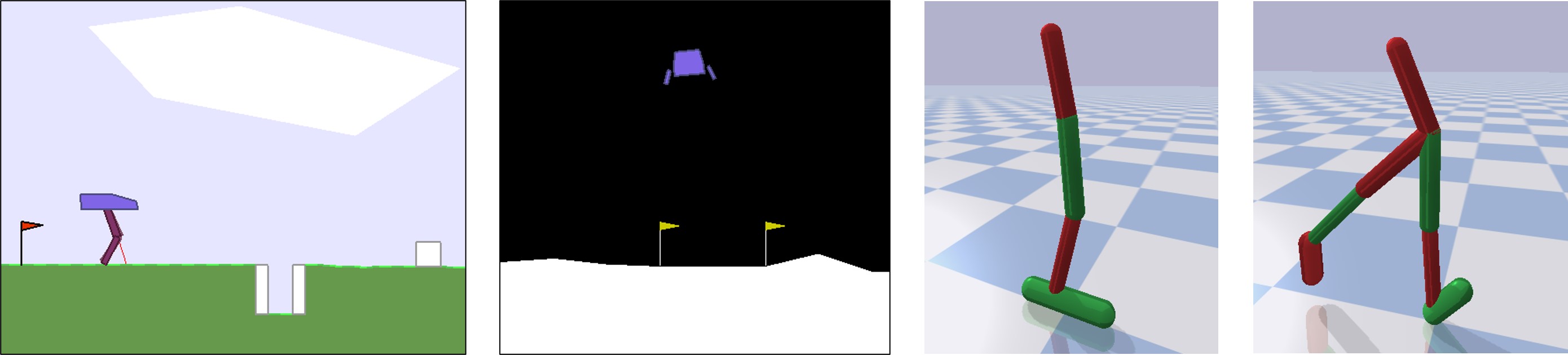}}
\caption{Screenshot of Benchmarks used in this paper (from left to right), BipedalWalker(Hardcore/System), LunarLander, Hopper, and Walker which tests agent's robustness against a variety of obstacle courses / actuator failure.}
\label{fig:env}
\end{center}
\vskip -0.2in

\end{figure}

\textbf{\textit{BipedalWalkerHardcore}}~\cite{brockman2016openai} involves agent observing the world with LIDAR, IMU, and joint encoder values to control torque on each of the bipedal walker's leg servos. The goal is to traverse through a randomly generated obstacle course filled with stairs, pitfalls, and walls. While individual obstacles are easy, the challenge is to learn a robust policy that can solve a variety of sequences of obstacles without falling.

\textbf{\textit{BipedalWalkerSystem}} is a modified version of the above where the agent traverses through a fixed sequence of obstacles with simulated random system failures. When a failure is triggered at a random timestep, the affected servo will only be able to deliver 60 -100\% of the original power depending on the severity. While individual scenarios are easy, the challenge is to learn a robust kinetic energy management skill to go through obstacles even if an actuator fails.

\textbf{\textit{LunarLander}} is a modified benchmark of the one provided by ~\cite{brockman2016openai}. 
Using position and velocity observations, an agent has to safely land on the landing pad using its main engine (ME) and side thrusters. When failure is triggered at a random timestep, the throttle of the affected rocket motor is limited to 60 - 100\% of the nominal power. Under nominal scenarios, the best policy is to wait until the last moment and fire the ME at full thrust to minimize fuel used. However, if a system failure occurs, the lander may crash due to ME being unable to provide enough thrust. A robust policy should keep the rate of descent to a manageable level to maximize the possibility of landing even when a failure occurs.  

\textbf{\textit{Hopper}} and \textbf{\textit{Walker}} are modified benchmarks based on the original versions provided by~\cite{benelot2018}. When a random system failure occurs, a torque limit of 75 - 100\% of the nominal maximum is applied to the affected servo. To make the benchmarks more challenging, we mount a simulated payload of sizes 0.75 and 0.5 on the Hopper and Walker legged robots.  A policy is considered to have failed if its simulated payload touches the ground.

\subsection{Baseline Algorithms}

Some of the state of the art approaches have been chosen as follows. To compare GC against adversarial RL approaches, we chose RARL ~\cite{pinto2017robust}, RARARL ~\cite{pan2019risk}, and PRMDP / NRMDP ~\cite{tessler2019action}. We chose FPO ~\cite{paul2018fingerprint} for comparison against approaches that control the training environment. For comparison against curricular RL, we chose SPDL ~\cite{klink2020self} for parametric curricular approaches and POET ~\cite{wang2019paired} for non-parametric approaches.

The algorithms in this paper require a base RL algorithm for updating policies. RL algorithms listed on top of the respective leaderboards for the original version of the benchmarks are used as base RL algorithms. BipedalWalker and LunarLander use soft actor-critic (SAC) ~\cite{liu}, while Hopper and Walker use twin delayed DDPG (TD3) ~\cite{stable-baselines3}. SAC uses $\gamma=0.99$, learning rate of 1e-4, batch size of 100, and replay memory size of 1e6, while TD3 uses $\gamma=0.98$, learning rate of 3e-4, batch size of 100, and replay memory size of 2e5. Both algorithms use fully connected networks consisting of layers sized 400 and 300 updated by ADAM ~\cite{kingma2014adam} and activated with the ReLU function.

\subsection{Evaluation and Hyperparameters}

One of the main challenges for comparing the performance of each algorithm is the vastly different computation requirements of each algorithm during training. FPO, POET, SPDL, and ours require additional steps for evaluating the agent's performance. While this can be expensive, evaluation can not only run in parallel but is also cheaper than exploration which requires backpropagation. Adversarial RL algorithms, on the other hand, had extra computational costs for training both protagonist and adversarial networks at the same time. As we are concerned about how robust a converged solution is, we report results based on how many epochs have passed. Each epoch consists of the same numbers of policy updates and exploration steps per benchmark. To share insights in cases where the total number of environment interactions is more important, we also include a separate set of experiments on the LunarLander benchmark where values are reported based on how many steps each algorithm interacted with the simulator. 

We report performance with mean and standard error on 10 random seeds per algorithm per benchmark, with testing sets consisting of randomly generated scenarios. For BipedalWalker benchmarks, only 3 random seeds were used to balance accuracy and computational cost. This is because it would take 7 - 14 days to approach convergence for such benchmarks. In the case of POET where multiple agents are trained simultaneously, we report the performance of the best agent in terms of reward as the result of the random seed. ~\Cref{tab:genetic_hyperparameter} shows the length of each epoch, testing set size, and the number of epochs for each benchmark.

\begin{table}
\centering
\caption{Training Duration and Testing Resolution}
\begin{tabular}{c||l|l|l}
 Benchmark & \makecell{Steps per \\ Epoch} & \makecell{Testing \\ Set Size} & \makecell{Number of \\ Epochs}  \\ \hline
 BipedalWalkerHardcore & 1e5 & 1000 & 350\\ \hline
 BipedalWalkersystem & 1e5 & 2500 & 30 \\ \hline
 LunarLander & 1e4 & 2500 & 80 \\ \hline
 Walker & 5e4 & 2500 & 60 \\ \hline
 Hopper & 5e4 & 2500 & 40
\end{tabular}
\label{tab:genetic_hyperparameter}
\vskip -2mm
\end{table}

To offer a fair comparison, a hyperparameter search is conducted for adversarial RL algorithms as shown in ~\Cref{tab:hyperparameter}. Hyperparameters that performed the best overall throughout the benchmarks were selected.

\begin{table}
\centering
\caption{Hyperparameter Search on Baseline Algorithms}
\begin{tabular}{c||l|l|l}
 Algorithm & Parameter & Tested & Selected \\ \hline
 RARAL & $\alpha$ & 0.05,0.1,0.5 & 0.1 \\ \hline
 RARARL & $\xi$ & 1,5,10,20 & 10 \\ \hline
 PRMDP & $\alpha$ & 0.05,0.1,0.3,0.5 & 0.1 \\ \hline
 NRMDP & $\alpha$ & 0.05,0.1,0.3,0.5 & 0.05 
\end{tabular}
\label{tab:hyperparameter}
\vskip -6mm
\end{table}

The size of policy evaluation for FPO, POET, SPDL is the same as the size of policy evaluation used for reporting performance during training. POET also requires manual reward thresholds on what is considered as not too trivial nor difficult before adding an scenario for training. BipedalWalker benchmarks use the same threshold of 50 - 300 as used in the original POET paper. For other benchmarks, we selected the value by checking their training curves and marking when the reward starts to climb and flatten. The threshold is set as 100 - 250 for LunarLander and 1000 - 2000 for Walker and Hopper benchmarks.

The default versions of the benchmarks random use number generators to create scenarios at each run. For FPO and SPDL, we engineered a fixed-length encoder where the range of the numbers coming out of the random number generator was defined. For POET and our method, the string of numbers to be used in the place of the random number generator was stored in a sequence.

For GC, the curriculum size is 300 for BipedalWalker benchmarks and 100 for the rest, which is a rounded value on how many times the simulators are reset per each epoch. The size of parent and offspring populations is 100 each which is a rounded value on the minimum size required to have at least two or three failure sequences upon random initialization to act as parents for subsequent generations. While hyperparameter tunning was also conducted on $p_{\mu}$, GC didn't show much sensitivity towards $p_{\mu}$ and a value of 0.1 is used. The GC's reliance on crossover more than the mutation rate is highlighted in the ablation study.

\subsection{Ablation Study}
 
To better understand how our approach help improves the robustness of an agent, an ablation study with BipedalWalkerHardcore is conducted. When generating a curriculum filled with failure scenarios, NoMutation turns off mutation, NoCrossover turns off crossover, and RandomFailure fills a curriculum with examples that are randomly generated and are unsolvable when tested by current policy. To see how a genetic algorithm can generate a curriculum of similar examples and its effect on performance, the mean genetic distance of a curriculum is also reported. Every time a new example is loaded during training, genetic distance is calculated by counting the minimum number of variables that have to be changed, added, and deleted to convert the previous example to the new example. Single Run provides additional data on the effect of genetic distance on robustness by generating a curriculum consisting of one failure scenario, making the mean genetic distance of the curriculum to be zero.

\section{Results}

\subsection{Comparison with Baseline Algorithms}

\begin{table*}[]
\centering
\caption{Reward and Mean Failure Rate of Trained Agents($\%$)}
\begin{tabular}{l||r|r|r|r|r}
\textbf{Reward}\\ \hline
 Algorithm&  BipedalWalkerHardcore&  BipedalWalkerSystem& LunarLander & Walker & Hopper \\ \hline
 Base RL (SAC / TD3) & 291.76 $\pm$ 17.41 & \textbf{300.94 $\pm$ 1.85} & 265.30 $\pm$ 1.92           & 2300.81 $\pm$ 29.30         & 2266.64 $\pm$ 3.05\\ \hline
 RARL     & 7.67 $\pm$ 13.49              & 289.25 $\pm$ 5.08          & 28.00 $\pm$ 12.24           & 122.60 $\pm$ 4.58           & 203.12 $\pm$ 4.53\\ \hline
 RARARL   & 230.14 $\pm$ 19.52            & 270.89 $\pm$ 13.59         & \textbf{272.29 $\pm$ 0.80}  & 2156.48 $\pm$ 10.31         & 2199.82 $\pm$ 3.18\\ \hline
 PRMDP    & 285.30 $\pm$ 25.66            & 298.42 $\pm$ 0.23          & 260.73 $\pm$ 2.28           & 2165.19 $\pm$ 11.36         & 2275.72 $\pm$ 2.04\\ \hline
 NRMDP    & 289.82 $\pm$ 19.25            & 291.42 $\pm$ 5.05          & 254.99 $\pm$ 1.27           & 2147.51 $\pm$ 1.30          & 2092.10 $\pm$ 3.85\\ \hline
 FPO      & 118.60 $\pm$ 1.21             & 286.47 $\pm$ 10.84         & 256.31 $\pm$ 4.61           & 2134.83 $\pm$ 5.09          & 2044.73 $\pm$ 3.21\\ \hline
 POET     & 24.60 $\pm$ 18.61             & -62.58 $\pm$ 14.14         & 213.30 $\pm$ 3.94           & 2068.50 $\pm$ 31.01         & 2129.93 $\pm$ 1.81\\ \hline
 SPDL     & \textbf{305.90 $\pm$ 0.45}    & 289.13 $\pm$ 5.19          & 221.31 $\pm$ 10.71          & 589.78 $\pm$ 74.04         & 2274.70 $\pm$ 13.06\\ \hline
 GC (Proposed)& \textbf{304.33 $\pm$ 1.65}& \textbf{300.00 $\pm$ 1.00 }& \textbf{272.82 $\pm$ 0.30} & \textbf{2342.61 $\pm$ 5.45}          & \textbf{2283.48 $\pm$ 2.01}\\ \hline
 
 \textbf{Failure Rate($\%$)}\\ \hline
 Algorithm&  BipedalWalkerHardcore&  BipedalWalkerSystem& LunarLander & Walker & Hopper \\ \hline
 Base RL (SAC / TD3)  & 10.20 $\pm$ 0.71       & 3.62 $\pm$ 0.58         & 5.19 $\pm$ 1.20          & 4.31 $\pm$ 1.00         & 12.99 $\pm$ 3.52\\ \hline
 RARL                 & 91.27 $\pm$ 3.28       & 5.97 $\pm$ 2.87         &73.56 $\pm$ 13.52         & 79.01 $\pm$ 16.82       & 85.61 $\pm$ 9.77\\ \hline
 RARARL               & 27.69 $\pm$ 3.14       & 15.29 $\pm$ 5.27        & 4.33 $\pm$ 0.88          & 3.85 $\pm$ 0.41         & 14.36 $\pm$ 5.27\\ \hline
 PRMDP                & 11.23 $\pm$ 0.36       & 2.85 $\pm$ 0.40         & 2.24 $\pm$ 0.90          & 3.88 $\pm$ 1.75         & 12.46 $\pm$ 4.70\\ \hline
 NRMDP                & 11.00 $\pm$ 1.27       & 5.02 $\pm$ 1.38         & 6.41 $\pm$ 1.15          & 6.96 $\pm$ 2.09         & 28.32 $\pm$ 5.91\\ \hline
 FPO                  & 67.60 $\pm$ 19.05      & 8.30 $\pm$ 4.55         &11.73 $\pm$ 2.71          & 12.12 $\pm$ 4.83        & 38.33 $\pm$ 5.11\\ \hline
 POET                 & 84.96 $\pm$ 9.45       & 100 $\pm$ 0.00          &29.53 $\pm$ 2.36          & 12.31 $\pm$ 7.40        & 24.98 $\pm$ 7.26\\ \hline
 SPDL                 & 20.87 $\pm$ 7.40       & 12.57 $\pm$ 2.41        &27.47 $\pm$ 4.13          & 21.18 $\pm$ 6.56        & 8.69 $\pm$ 6.57\\ \hline
 GC (Proposed) & \textbf{3.96 $\pm$ 0.37}      & \textbf{2.16 $\pm$ 0.45}& \textbf{0.64 $\pm$ 0.02 }& \textbf{2.35 $\pm$ 1.11}&\textbf{7.30 $\pm$ 2.79}\\ \hline
\end{tabular}
\label{tab:results}
\vskip -5mm
\end{table*}

\vspace{-0.8cm}
\begin{figure}[ht]
\vskip 0.2in
\begin{center}
\centerline{\includegraphics[width=0.9\columnwidth]{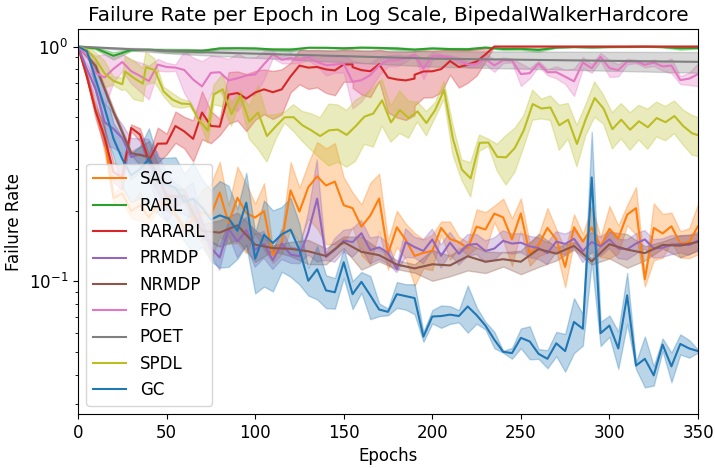}}
\caption{Training Curve for BipedalWalkerHardcore. Lower the better}
\label{fig:learning}
\end{center}
\vskip -0.2in
\end{figure}

As shown in ~\Cref{fig:learning} and ~\Cref{tab:results}, our proposed GC consistently improves over the state of the art algorithms especially in terms of robustness. Agents trained by GC are 2 - 8x times less likely to fail compared to those trained on baseline SAC / TD3.

One interesting observation from ~\Cref{tab:results} is that even when agents show quite a difference in performance in terms of robustness, such differences are less obvious when looking at rewards only. When trained, big reward coming from the majority of the cases tends to dominate over bad rewards coming from the minority of cases. During training, similar issues are observed where the failure rate continues to converge when the reward does not show significant progress. This highlights the challenge of capturing robustness alone by reward and using it to optimize the policy. 

The results from FPO and SPDL highlight the GC's benefit of using raw scenario encoding of non-fixed length. As a non-parametric optimizer, GC can adapt the length of its encoding to generate more complex scenarios as the protagonist agent's performance improves. The effect is most well observed in the case of BipedalWalkerHardcore where GC expanded the encoding size by 10 - 15 times during training to precisely describe the what training scenarios should be.

Also, the results from adversarial RL show selecting a supporting curriculum is as important as generating challenging scenarios during training. While adversarial agents can generate challenging situations during training, they do not present in a way that the trainee can easily learn new skills. The challenges of injecting difficult cases without supporting a curriculum are further highlighted in the following section on the ablation study. 

The effect of not nesting training within the genetic algorithm can be observed from POET's results. The original POET paper reported a reward of around 250 when trained and tested on scenarios based on the BipedalWalkerHardcore. POET showed a similar performance when evaluated on the training set during our experiments. However, when evaluated against the testing set which includes the entire scenario space as designed by the benchmark, POET performed poorly agent's skills were not generalizable across a wide variety of scenarios. As POET is computationally expensive and relies only on mutation for curriculum generation, each trained agent could only experience less training data from a less diverse set of scenarios compared to those from GC.

An interesting observation from the trained policies is while there are some scenarios where an agent had more difficulty solving than the others, there was no clear trend describing which scenario is objectively more difficult than others or a priori difficult scenarios where solving one case means being able to solve all the easier cases. When a scenario that a trained agent consistently fails are used as a training scenario to a randomly initialized agent, the agent would learn how to solve the scenario. However, regardless of the random seed used, finding a general policy that solves all the scenarios was difficult. This shares insight that a robust training scheme should not only focus on performance per each task but also on learning a general skillset applicable across the tasks.

\subsection{Comparison with Respect to Environment Interactions}

\begin{figure}[ht]
\begin{center}
\centerline{\includegraphics[width=0.9\columnwidth]{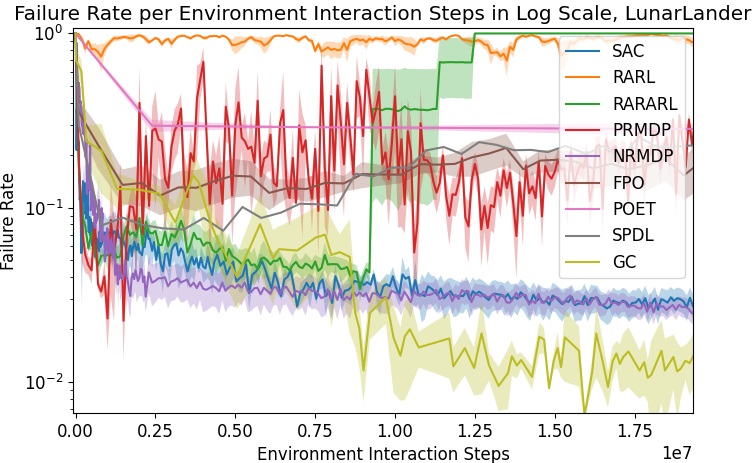}}
\caption{Characteristic Training Curve from LunarLander Benchmark with Respect to Environment Interaction Steps}
\label{fig:compcost}
\end{center}
\vskip -0.2in
\end{figure}

One of the important criteria in RL is how efficiently it can learn per the number of environment interactions. ~\Cref{fig:compcost} shows that while GC is a bit slow at the start due to the extra cost of running genetic algorithms, the cost is offset by having better training examples. Unlike the baseline RL (SAC) and adversarial RL methods where marginal utility per environment interaction quickly diminishes, GC can sustain the rate of performance improvement longer and converges to a better solution.

\subsection{Ablation Study}

\begin{table}[]
\caption{Reward, Failure Rate, and Mean Genetic Distance between Training Examples during Ablation Study}
\begin{tabular}{l|l|l|l}
 Method        & Reward & Failure Rate(\%)  & Genetic Distance \\ \hline
 Base RL (SAC) & 291.76 & 10.2 &  22.65\\ \hline
 GC (Ours)     & \textbf{304.33} & \textbf{3.96} & \textbf{10.60} \\ \hline
 No Mutate     & 294.17          & 8.51          & \textbf{10.44} \\ \hline
 No Crossover  & 271.72          & 17.63         & 20.92 \\ \hline
 Random Failure& 251.37          & 24.50         & 23.34 \\ \hline
 Single Run    & 99.45           & 33.33         & 0
\end{tabular}
\label{tab:ablation}

\end{table}

\begin{figure}[ht]
\vskip 0.2in
\begin{center}
\centerline{\includegraphics[width=0.9\columnwidth]{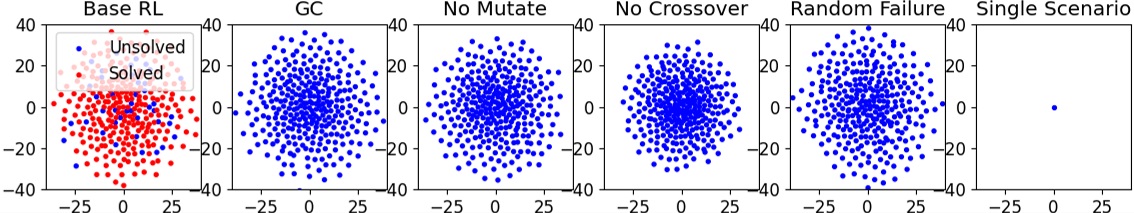}}
\caption{tSNE analysis on genetic distance of generated scenarios}
\label{fig:tSNE}
\end{center}
\vskip -0.2in
\vspace{-0.3cm}
\end{figure}

One of the insights from ~\Cref{tab:ablation} is, except for Single Run, curricula with similar scenarios, i.e. shorter mean genetic distance, perform better. While Random Failure builds a curriculum with failed scenarios, the similarity between the scenarios is not ensured. In the case of No Crossover, genetic similarity between scenarios is low as unlike crossover which mixes sequences from two parents to generate two offsprings, mutation only creates one offspring from one parent. The difficulty in transferring skills between more distant scenarios seems to result in No Crossover and Random Failure performs poorly.

~\Cref{fig:tSNE} on the other hand highlights how a coverage over scenario space affects performance for curriculums with short mean genetic distance between scenarios. As an agent is trained based on scenarios it experiences, having curriculum scenarios more spread out in the scenario space can help the agent generalize across diverse scenarios. While Single Run keeps genetic distance between scenarios to a minimum, it offers a poor coverage of the scenario space as in ~\Cref{fig:tSNE}. Genetic Distance between scenarios generated by No Mutate is similar to those in GC, but the former offers narrower coverage of the scenario space. While No Mutate can only reorganize genetic sequences it had upon initialization, GC can introduce new sequences through mutation, allowing it to explore a wider scenario space and train a more robust agent. 

\section{Conclusion and Future Works}

This paper proposes genetic curriculum, an RL algorithm that uses a genetic algorithm to generate a curriculum of scenario for training RL agents. Through empirical study, our algorithms show improvement over existing state-of-the-art approaches concerning robustness. Future works will focus on decreasing the computational load of our algorithms, improving rate of convergence, as well as implementing our method in real and more complex benchmarks.



\bibliographystyle{ieeetr}
\bibliography{references}

\end{document}